ICDSST 2020 PROCEEDINGS – ONLINE VERSION
THE EWG-DSS 2020 INTERNATIONAL CONFERENCE ON DECISION SUPPORT SYSTEM TECHNOLOGY
I. Linden, M.T. Escobar, A. Turón, F. Dargam, U. Jayawickrama (editors)
Zaragoza, Spain, 27-29 May 2020

# ICDSST 2020 on
# Cognitive Decision Support Systems & Technologies

# Group Decision Support for agriculture planning by a combination of Mathematical Model and Collaborative Tool

Pascale Zaraté[1][0000-0002-5188-1616], MME Alemany[2][0000-0002-0992-8441], Ana Esteso Alvarez[2][0000-0003-0379-8786], Amir Sakka[1,3] and Guy Camilleri[1][0000-0003-2916-4589]

[1]IRIT, Toulouse University, Toulouse, France
[2]CIGIP, UniversitatPolitècnica de València, Camino de Vera S/N, 46002 Valencia, Spain
[3] IRSTEA, Clermont-Ferrand, France
{zarate,camiller,sakka}@irit.fr
mareva@omp.upv.es
aesteso@cigip.upv.es

## ABSTRACT

Decision making in the Agriculture domain can be a complex task. The land area allocated to each crop should be fixed every season according to several parameters: prices, demand, harvesting periods, seeds, ground, season etc… The decision to make becomes more difficult when a group of farmers must fix the price and all parameters all together. Generally, optimization models are useful for farmers to find no dominated solutions, but it remains difficult if the farmers have to agree on one solution. We combine two approaches in order to support a group of farmers engaged in this kind of decision making process. We firstly generate a set of no dominated solutions thanks to a centralized optimization model. Based on this set of solution we then used a Group Decision Support System called GRUS for choosing the best solution for the group of farmers. The combined approach allows us to determine the best solution for the group in a consensual way. This combination of approaches is very innovative for the Agriculture. This approach has been tested in laboratory in a previous work. In the current work the same experiment has been conducted with real business (farmers) in order to benefit from their expertise. The two experiments are compared.

## INTRODUCTION

Each season farmers must face the difficult decision about which crops to be planted and the allocated land area to each of them. Farmers usually make this decision based on market prices of the crops in the previous season. However, market prices highly depend on the balance between supply and demand. In this context, if most of farmers decide to cultivate the crops more profitable the previous year, an excess of these crops could provoke both, a decrease in their prices and high quantities of waste. The imbalance between demand and supply is largely due to the lack of collaboration among farmers who individually decide about planting and harvesting decisions.

With the aim of supporting farmers in this difficult task, centralized mathematical programming models integrating decisions on planting and/or harvesting for a set of farmers have been developed that provide the optimal solution for the agricultural supply chain. However, this centralized approach could produce inequalities in the profits obtained by farmers, leading to the unwillingness to cooperate and contribute to the collaborative crop planning, and to the farmers unacceptance of the obtained planning. This task is even more difficult when several objectives are taken into account. In this context, approaches exist that can be used to find a predetermined number of non-dominated solutions. However, the problem even remains difficult if farmers have to agree on one solution to be implemented.

In order to support the farmers for this complex task we combine two approaches. We firstly generate a set of non-dominated solutions based on a centralized optimization model. The number of non-dominated solutions should be defined in advanced by the group of farmers. Once decided and based on the set of non-dominated solution obtained, we then used a Group Decision Support System called GRUS for choosing the best solution for the group of farmers. The combined approach allows us to determine the best solution for the group in a consensual way. This combination of approaches is very innovative for the Agriculture domain.

In [1], we combined these two approaches in order to generate a satisfactory solution for a group of human beings. First of all, we generated 10 solutions thanks to a centralized optimization model. These solutions are then explained to a group of five end-users playing the role of farmers. We, in a second step, asked to the five end-users to give their own preferences on these 10 solutions using a Group Decision Support System (GDSS), called GRUS. This GDSS allows to find the final ranking for the group based on the preferences given by the stakeholders.. It was shown in this study how the GDSS GRUS is helpful to generate a group decision which reduces conflicts in a group and how it supports to find a consensus. In the GDSS we used the Borda voting procedure [2]. Nevertheless, the conclusions of this experiment have some limitations based on the fact the decision makers were researchers and not farmers. In this current paper, we conducted the same experiment with real agriculture businessmen. The main objective of this paper is to compare the results found in laboratory with results obtained with real users.

The paper is structured as follows. First, the problem under study is described. Second, the methodology to generate multiple solutions. Then, the group decision procedure to select one of them in a laboratory and real contexts and their comparison are presented. Finally, a set of conclusions are derived.

**THE PROBLEM**

The agricultural supply chain under study is one typical of the region of La Plata, in Argentina, for the tomato crop. The supply chain is assumed to be integrated by different farmers that are in charge of the production, cultivation, harvesting and distributing of different varieties of tomatoes to several markets. A mixed integer linear programming model (MILP) was developed to support the centralized decision making about the planting and harvesting decisions per farmer and tomato variety, the quantity of each type of tomato to betransported from the farmer to each market, the waste as well as the unfulfilled demand.Five farmers were considered with an available planting area in hectare (ha) for each farmer of 20, 18, 17, 16 and 15, respectively. The planning horizon was one year divided into months. Three tomato varieties were considered: pear, round and cherry. The planting period comprises three different months (July, October, and January) that do not depend on the specific variety. The harvesting periods and the yield are dependent on the planting period but are also the same for all the tomato varieties (Table 1). These planting periods are the usual in the region of La Plata, that is one of the most important areas of tomato in greenhouse for sell in fresh in Argentina.

**Table 1.** Harvesting periods

|         | Jul | Aug | Sep | Oct | Nov | Dec | Jan | Feb | Mar | Apr | May | Jun |
|---------|-----|-----|-----|-----|-----|-----|-----|-----|-----|-----|-----|-----|
| July    |     |     |     |     | X   | X   | X   | X   |     |     |     |     |
| October |     |     |     |     |     |     | X   | X   | X   | X   |     |     |
| January |     |     |     |     |     |     |     |     | X   | X   | X   | X   |

From the planted date to the harvesting date, different activities need to be made to the plant in order to ensure its correct growth. Because the tomato crop matures over time, it is necessary to make different harvesting passes in the same time period whose time per plant depends on the tomato variety. Both activities require of human labor that is assumed to present limited capacity.

Once harvested the tomatoes are distributed to two different customers, a central market and some restaurants, incurring in a transport cost that depends on the farmer and the customer. The transported quantities to each market try to satisfy the monthly demand of variety taking into account the sale prices that is dependent on the ration between supply and demand. Lack of supply to cover market demand is modelled by the decision variable of unmet demand and excess of supply as regards market demand is reflected by the waste decision variables.

The decisions made should respect limitations about the available planting area in each farm and other supply chain policies related to the planting of tomato varieties: all tomato varieties should be planted in all planting periods and all farmers should plant some variety in all planting periods. Other constraints reflected the balance equations between quantities planted and harvested and these last ones with transported quantities and fulfilled demand. The waste in each farm is calculated as the difference between matured tomatoes and those not harvested meanwhile the waste in markets as the difference between quantities transported and not sold.

When searching for a solution the three dimensions of sustainability are taken into account by incorporating them into the multi-objective model. The three objectives considered were:
- Economic Objective: Maximize the profits of the supply chain as the difference between incomes per sales and total costs.
- Environmental Objective: Minimize the total waste along the Supply Chain.
- Social Objective: Minimize the unfulfilled demand along all the Supply Chain covering human requirements and increasing the customer satisfaction.

In its current state, the experiment does not take into account the fact that side payments would be possible to make the generated solution acceptable for all group members. Instead, the GDSS GRUS is used to choose the most satisfactory solution for the group between non-dominated solutions whose generation is reported in the following section.

**GENERATING NON-DOMINATED SOLUTIONS FOR THE MULTI-FARMER PLANTING MODEL**

To solve the centralised multi-objective mixed integer linear programming model, we transformed it into a single-objective model by applying the ε-constraint method ([3]; [4]).The method starts optimizing the model only for one objective. The optimal value of this first objective is used to formulate a constraint for the next model execution that in this step is optimized for a second objective. The same process is made with the third objective by constraining both the first and second objective. The process is repeated now starting from another objective and so on until all the different combinations of the objectives are solved. This provides with a set of solutions from which dominated solutions are discarded. The non-dominated solutions are analyzed to identify the best and worst values for each objective that provide the range of values used to define the grid points for which the model will be solved.

For our case study, ten values were defined for the $\varepsilon_i$ parameter. The model was implemented using the MPL software 5.0.6.114 and the solver Gurobi 8.0.1. This provide us with ten non-dominated solutions. The detail for each non-dominated solution can be consulted in Figure 1. For each solution, the value of the three objective functions for the entire supply chain are presented. Readers are referred to [1] to consult the solution for each farmer. It is necessary to find a complementary procedure to decide which non-dominated solution to implement for two reasons: 1) because of being non-dominated solutions the profit, wastes and unfulfilled demand that reports the best result for one objective is not the best for the others and 2) there are multiple farmers affected that will not base their decisions only on SC objectives but also and mainly in their own objectives. This procedure based on the GRUS system is described in the following section.

| Solution | Name | SC Profits (€) | SC wastes (kg) | Unfulfiled demand (kg) |
|---|---|---|---|---|
| 1 | A | 148.334.625 | 5.316.020 | 207.317.999 |
| 2 | B | 148.302.280 | 5.315.998 | 201.749.612 |
| 3 | C | 148.003.481 | 6.417.520 | 195.841.392 |
| 4 | D | 146.849.751 | 11.193.326 | 189.933.239 |
| 5 | E | 145.326.260 | 14.017.213 | 184.025.050 |
| 6 | F | 142.518.888 | 11.213.768 | 178.116.854 |
| 7 | G | 136.863.913 | 8.410.330 | 172.208.666 |
| 8 | H | 146.572.577 | - | 204.769.167 |
| 9 | I | 135.083.010 | - | 182.724.221 |
| 10 | J | 129.129.328 | 25.230.996 | 154.484.078 |

Figure 1. Objective values for the non-dominated solutions generated

**GRUS RESEARCH EXPERIMENT USING CENTRALIZED MODEL SOLUTIONS**

We used GRUS to rank the 10 generated solutions for the multi-farmer planting model which anticipates harvesting and transporting decisions. We had five decision makers playing the role of the farmers, including the facilitator as a decision maker. This experiment was conducted in research laboratory. The adopted process was composed by the following three steps:
1. Alternatives generation: The facilitator filled in the system the 10 non-dominated solutions found thanks to the optimization model.
2. Vote: The five decision makers ranked the 10 solutions according to their own preferences. For this, the value of each objective jointly with the area planted with each tomato variety for each farmer and for the whole supply chain were provided to decision-makers.
3. Ranking solutions: The system then computes the final ranking for the group using the Borda [2] methodology.

The results of this procedure are described in the Fig. 2.

| Research Experiment (RE) | | |
|---|---|---|
| Solution | Points | Ranking |
| A | 17 | 4 ex aequo |
| B | 20 | 3 |
| C | 23 | 2 |
| D | 24 | 1 |
| E | 17 | 4 ex aequo |
| F | 16 | 5 ex aequo |
| G | 16 | 7 |
| H | 10 | 5 ex aequo |
| I | 15 | 6 |
| J | 8 | 8 |

**Fig. 2.** Result of the Group Ranking provided by GRUS for the research experiment.

This result is given for the group of five end-users. The five farmers have the same weight (importance) for this experiment. Nevertheless, we also could choose that the importance of each farmer is linked to the number of hectares, only in Multi-Criteria processes.

We can see that on positions 4 and 5 two alternatives are ex aequo: solutions A and E for rank 4 and solutions F and H for rank 5. The best solution for the group is the one for which the five farmers have benefits and the three tomato varieties are planted, that is solution D. Nevertheless, we can notice that it is not the solution, which generates the best profit on a global point of view.

This first experiment shows that the solution obtained by a centralized optimization model that generates the highest profit, that is the solution A, is not necessarily the best one for the group of agents (humans). In order to show that this combination of approaches could be useful in real situations, we conducted again the same experiment with real businessmen in agriculture (farmers).

**GRUS BUSINESS EXPERIMENT USING CENTRALIZED MODEL SOLUTIONS**

We again used GRUS to rank the same 10 generated non-dominated solutions. Four decision makers who were businessmen in agriculture, including the facilitator as a decision maker, gave their own preferences. The same process composed by the above three steps was adopted but the business had a higher importance than the facilitator. For this second experiment the five end-users did not have the same importance. The facilitator had the lowest importance (1) and two of the businessmen had the highest importance (5). The two other businessmen had a medium importance (3). The results can be consulted in Fig. 3. We can observe that for businessmen selection, the solution with the highest profit is the best in the new ranking.

| Business Experiment (BE) | | |
|---|---|---|
| Solution | Points | Ranking |
| A | 30 | 1 |
| B | 26 | 4 |
| C | 17 | 7 ex aequo |
| D | 18 | 6 |
| E | 11 | 8 |
| F | 17 | 7 ex aequo |
| G | 28 | 3 |
| H | 20 | 5 |
| I | 29 | 2 |
| J | 6 | 9 |

**Fig. 3.** Result of the Group Ranking provided by GRUS for the business experiment

The comparison of the final ranking between the business experiment (BE) and the research experiment (RE) can be consulted in Fig. 4, where the last three columns on the right represents the difference for each objective between solutions for businessmen and researchers ranked in the same position. A positive difference for SC profits means that businessmen solution is better in profit than researcher solution because the objective is to maximize profits. On the contrary a negative difference in SC wastes and unfulfilled demand represents a better solution for these two objectives for businessmen since the objective is to minimize them.

| | | | Business Experiment (BE) | | | | | | Research Experiment (RE) | | | DIFFERENCES= BE-RE | | |
|---|---|---|---|---|---|---|---|---|---|---|---|---|---|---|
| Solution | SC Profits (€) | SC wastes (kg) | Unfulfilled demand (kg) | Ranking | Points | Solution | SC Profits (€) | SC wastes (kg) | Unfulfilled demand (kg) | Ranking | Points | SC Profits (€) | SC wastes (kg) | Unfulfilled demand (kg) |
| A | 148.334.625 | 5.316.020 | 207.317.999 | 1 | 30 | D | 146.849.751 | 11.193.326 | 189.933.239 | 1 | 24 | 1.484.874 | -5.877.306 | 17.384.760 |
| I | 135.083.010 | 0 | 182.724.221 | 2 | 29 | C | 148.003.481 | 6.417.520 | 195.841.392 | 2 | 23 | -12.920.471 | -6.417.520 | -13.117.171 |
| G | 136.863.913 | 8.410.330 | 172.208.666 | 3 | 28 | B | 148.302.280 | 5.315.998 | 201.749.612 | 3 | 20 | -11.438.367 | 3.094.332 | -29.540.946 |
| B | 148.302.280 | 5.315.998 | 201.749.612 | 4 | 26 | A | 148.334.625 | 5.316.020 | 207.317.999 | 4 ex aequo | 17 | -32.345 | -22 | -5.568.387 |
| H | 146.572.577 | 0 | 204.769.167 | 5 | 20 | E | 145.326.260 | 14.017.213 | 184.025.050 | 4 ex aequo | 17 | 1.246.317 | -14.017.213 | 20.744.117 |
| D | 146.849.751 | 11.193.326 | 189.933.239 | 6 | 18 | F | 142.518.888 | 11.213.768 | 178.116.854 | 5 ex aequo | 16 | 4.330.863 | -20.442 | 11.816.385 |
| C | 148.003.481 | 6.417.520 | 195.841.392 | 7 ex aequo | 17 | H | 146.572.577 | 0 | 204.769.167 | 5 ex aequo | 10 | 1.430.904 | 6.417.520 | -8.927.775 |
| F | 142.518.888 | 11.213.768 | 178.116.854 | 7 ex aequo | 17 | I | 135.083.010 | 0 | 182.724.221 | 6 | 15 | 7.435.878 | 11.213.768 | -4.607.367 |
| E | 145.326.260 | 14.017.213 | 184.025.050 | 8 | 11 | G | 136.863.913 | 8.410.330 | 172.208.666 | 7 | 16 | 8.462.347 | 5.606.883 | 11.816.384 |
| J | 129.129.328 | 25.230.996 | 154.484.078 | 9 | 6 | J | 129.129.328 | 25.230.996 | 154.484.078 | 8 | 8 | 0 | 0 | 0 |

**Fig. 4.** Comparison of the Group Ranking between the business experiment (BE) and the research experiment (RE)

Based on this, it can be observed in Fig. 4 that the rankings obtained for each group of decision-makers are different. It might seem that for businessmen is more important the SC profits than the other objectives because the solution A is the 1$^{rst}$ in their ranking. However, if we compared the 2$^{nd}$, 3$^{rd}$ and 4$^{th}$ solutions for the BE and RE, it can be observed that businessmen also consider important to minimize wastes and unfulfilled demand even worsening the SC profits. In any case, the alternative with the lowest profit and maximum waste stays the worst in the two rankings (J). It seems that for them the SC Profits and Wastes criteria are much more important than the SC unfilled demand criterion.

Comparing the two rankings, it can give a good representation on the weights of criteria for real decision makers. The difference between the research and the businessmen experiments ranking can be explained by the expertise that the businessmen have on how to manage this decision problem. Therefore, it can be concluded that the best solution for a group even based on quantitative objectives depend on the subjects involved being even more difficult to predict when multiple objectives should be considered.

**CONCLUSIONS**

Supporting a group of decision makers engaged in a decision process is generally a complex situation. Several stakeholders may involve conflicting situations that cannot be avoided. Two main difficulties can arise for a group of decision makers: determine the alternatives and determine the weight of criteria. Each decision makers can have his own preferences for the weight of criteria but determining the weight for the whole group of decision makers is a complex task considering that some decision makers could have more experience or have more importance in the group. In order to solve these issues, we combine two approaches. We firstly generate a set of non-dominated solutions thanks to solve a multi-objective centralized optimization model by means the ϵ-constraint method. The advantage of this automatic generation of solutions is that the decision makers can have a reasoning based on shared alternatives. Based on this set of solution, we then used a Group Decision Support System called GRUS for choosing the best solution for the group of farmers. The combined approach allows us to determine the best solution (or the least bad) for the whole group in a consensual way. This combination of approaches is very innovative for the Agriculture. The experiment was conducted twice: firstly, in a laboratory and then with real businessmen (farmers). As a result of these two experiments we can conclude that the weight of all used criteria is not the same for businessmen and researchers.

As perspective of this work and in order to avoid boring tasks to businessmen, like for example evaluate the weight of criteria, it would be interesting to calculate the weight of criteria by comparison with two experiments.

**ACKNOWLEDGMENTS**

The authors acknowledge the Project 691249, RUC-APS: Enhancing and implementing Knowledge based ICT solutions within high Risk and Uncertain Conditions for Agriculture Production Systems, funded by the EU under its funding scheme H2020-MSCA-RISE-2015. One of the authors acknowledges the partial support of the Programme of Formation of University Professors of the Spanish Ministry of Education, Culture, and Sport (FPU15/03595).